\pdfoutput=1

\documentclass[11pt]{article}
\usepackage[colorlinks = true,
            linkcolor = blue,
            urlcolor  = blue,
            citecolor = blue,
            anchorcolor = blue]{hyperref}

\usepackage[final]{acl}
\usepackage{booktabs}
\usepackage{times}
\usepackage{latexsym}
 \usepackage{float}
\usepackage[T1]{fontenc}

\usepackage[utf8]{inputenc}

\usepackage{microtype}

\usepackage{inconsolata}
\usepackage{amsmath}
\usepackage{amsfonts}
\usepackage{graphicx}
\usepackage{algorithm}
\usepackage{algorithmic}
\title{Whitening Not Recommended for Classification Tasks in LLMs}

\author{Ali Forooghi \\
  School of computer science\\
  University of Windsor\\ ON, Canada  \\
  \texttt{foroogh@uwindsor.ca} \\\And
  Shaghayegh Sadeghi \\
  School of computer science\\
   University of Windsor\\
    ON, Canada  \\
  \texttt{sadeghi3@uwindsor.ca} \\\And
   Jianguo Lu \\
  School of computer science \\
  University of Windsor\\
   ON, Canada  \\
  \texttt{jlu@uwindsor.ca} \\}

\begin{document}
\maketitle
\begin{abstract}
Sentence embedding is a cornerstone in NLP. Whitening has been claimed to be an effective operation to improve embedding quality obtained from Large Language Models (LLMs). However, we find that the efficacy of whitening is model-dependent and task-dependent. In particular, whitening degenerates embeddings for classification tasks.  The conclusion is supported by extensive experiments. 
A by-product of our research is embedding evaluation platform for LLMs called SentEval+ \footnote{\href{https://github.com/nlp-lab-dr-lu/senteval-plus}{\textcolor{blue}{Here is the link to the Github for SentEval+}}}
\end{abstract}

\section{Introduction}
Sentence embedding plays a fundamental role in NLP \cite{le2014distributed}. 
Despite the widespread success of Large Language Models (LLMs) in generative tasks, embeddings obtained from pre-trained models are not impressive
\cite{li2023angle}. Sometimes, they are not even competitive with traditional word2vec-based approaches on machine learning tasks such as classification and Semantic Text Similarity (STS). Consequently, there has been a flurry of research aimed at improving the quality of embeddings from pre-trained models \cite{gao2021simcse,jiang2022promptbert,li2023angle}.

Among this group of work, whitening has been shown to be an effective post-processing method for improving embeddings obtained from LLMs \cite{zhuo2023whitenedcse,su2021whitening,huang2021whiteningbert}. We find that the efficacy of whitening is both model-dependent and task-dependent. Although we reproduced the result that whitening does work for some models on STS tasks, it does not work for other models.  More importantly,  the effectiveness of the whitening operation is restricted to STS tasks. For classification tasks, whitening degrades embedding quality consistently and sometimes with a large margin. The result is supported consistently for all the evaluated models and all the datasets in SentEval  \cite{conneau2018senteval}.
 To further consolidate the surprising results, we explored a variety of whitening operations, including Principal Component Analysis (PCA) \cite{friedman1987exploratory}, Cholesky matrix decomposition \cite{siarohin2018whitening}, and Zero-Phase Component Analysis (ZCA) \cite{bell1997independent}. Although some variants of whitening induce different performances, the overall conclusion remains unchanged.

A by-product of our research is an embedding evaluation platform for LLMs, which we call SentEval$^+$,  to streamline the evaluation of embedding quality. LLMs are big and costly to run. SentEval \cite{conneau2018senteval} provides a platform for embedding evaluation on a variety of models, tasks, and datasets. It works well on smaller models such as BERT. To facilitate the evaluation of LLMs on commodity machines, we provide the embeddings for all sentences in our evaluation datasets.

There is not much detailed comparison of the performance of embeddings from OpenAI, maybe partially due to the cost for API calls. We observe that embeddings from OpenAI are on par with LLaMA overall.  Another interesting observation is that LLaMA and LLaMA2 are very close in terms of embedding performance. 
 
Our work is important for both practitioners and researchers in LLMs. For LLM providers such as openAI, various post-processing are commonly applied to the embeddings they serve. They may want to serve different types of embeddings for different tasks, with the understanding of our result. For researchers in the area,  running on a variety of LLMs is prohibitive computationally. Our SentEval$^+$ makes experiments feasible on commodity machines.

\section{Whitening Transformations}

LLM embeddings have the isotropy problem \cite{timkey2021all,kovaleva2021bert,rudman2021isoscore}. Whitening is a post-processing technique that converts spatially correlated, anisotropic feature representations into uncorrelated, isotropic ones \cite{sasaki2023examining,rudman2023stable}. For this purpose, whitening transforms the feature representations such that the mean is centred at the origin, covariances are eliminated, and the variance is normalized to an identity matrix. 

Given $N$ number of sentence embeddings $x_1, x_2, \dots,  x_N$.
Let $X=(x_1, x_2, \dots, x_N)^T  \in \mathbb{R}^{N \times d}$, where $d$ is the dimension of the embeddings.  The covariance matrix  for $X$ is $\Sigma = (X-\mu)(X-\mu)^T$,  where $\mu$ is the mean of $\{x_i\}_{i=1}^N$. 
Whitening transformation is achieved using a matrix $W$resulting in unit diagonal “white” covariance $var(Z) = I$:
\begin{align}
  Z&=W(X-\mu)   \label{zw} \\
W&= \begin{cases}
         U \Lambda^{-\frac{1}{2}}   &PCA\\
         U \Lambda^{-\frac{1}{2}} U^T   &{ZCA} \\
         L^T & Chol\\
        V \Theta^{-\frac{1}{2}} V^T & ZCA-Cor \\
        V \Theta^{-\frac{1}{2}}  & PCA-Cor
        \end{cases}  \label{eq:whitening}
\end{align}
$W$  in Equation \ref{zw}  varies as in Equation \ref{eq:whitening}.  The most commonly used whitening operation is called PCA-whitening, which is also the one used in the first a few papers on the performance gain of whitening on LLMs. Since our initial result on PCA-whitening shows the opposite for classification tasks, and \cite{wang2023investigating} reported different behaviour of ZCA-whitening, we exhaustively investigate all variations of whitening operations. 

In Equation \ref{eq:whitening},  $\Lambda$ is the eigenvectors, and $U$ is the eigenvalues of the covariance matrix, i.e.,  $\Sigma = U \Lambda U^T$.  The matrix $L$ corresponds to the Cholesky decomposition of the inverse of $\Sigma$, such that $LL^T = \Sigma^{-1}$. The matrices $V$ and $\Theta$ result from the eigen decomposition of the correlation matrix $P$, expressed as $P = V \Theta^{-\frac{1}{2}} V^T$, where $V$ is the eigenvector matrix and $\Theta$ contains the corresponding eigenvalues.

\begin{algorithm}[] 
            \caption{Whitening Operations}\label{alg:whitening}
            \begin{algorithmic}[1]
                \STATE \textbf{Input:} Embeddings $\{x_i\}_{i=1}^N$
                \STATE \textbf{Output:} Transformed embeddings $\{\tilde{x}_i\}_{i=1}^N$
                \STATE Compute the mean $\mu$ of $\{x_i\}_{i=1}^N$
                \STATE Compute the covariance matrix $\Sigma$ of $\{x_i\}_{i=1}^N$
                \STATE Compute the correlation matrix $P$ of $\{x_i\}_{i=1}^N$
                \STATE Let $U, \Lambda, U^T = \text{SVD}(\Sigma)$
                \STATE Let $V, \Theta, V^T = \text{SVD}(P)$
                \STATE Perform $LL^T = \text{Chol}(\Sigma^{-1})$
                \STATE Transform $\tilde{x}_i = (x_i - \mu)W$ using Eq. \ref{eq:whitening}
            \end{algorithmic}
\end{algorithm}             


\section{Experiments}

\begin{figure*}[ht]
    \centering
    
    \includegraphics[width=0.8\textwidth]{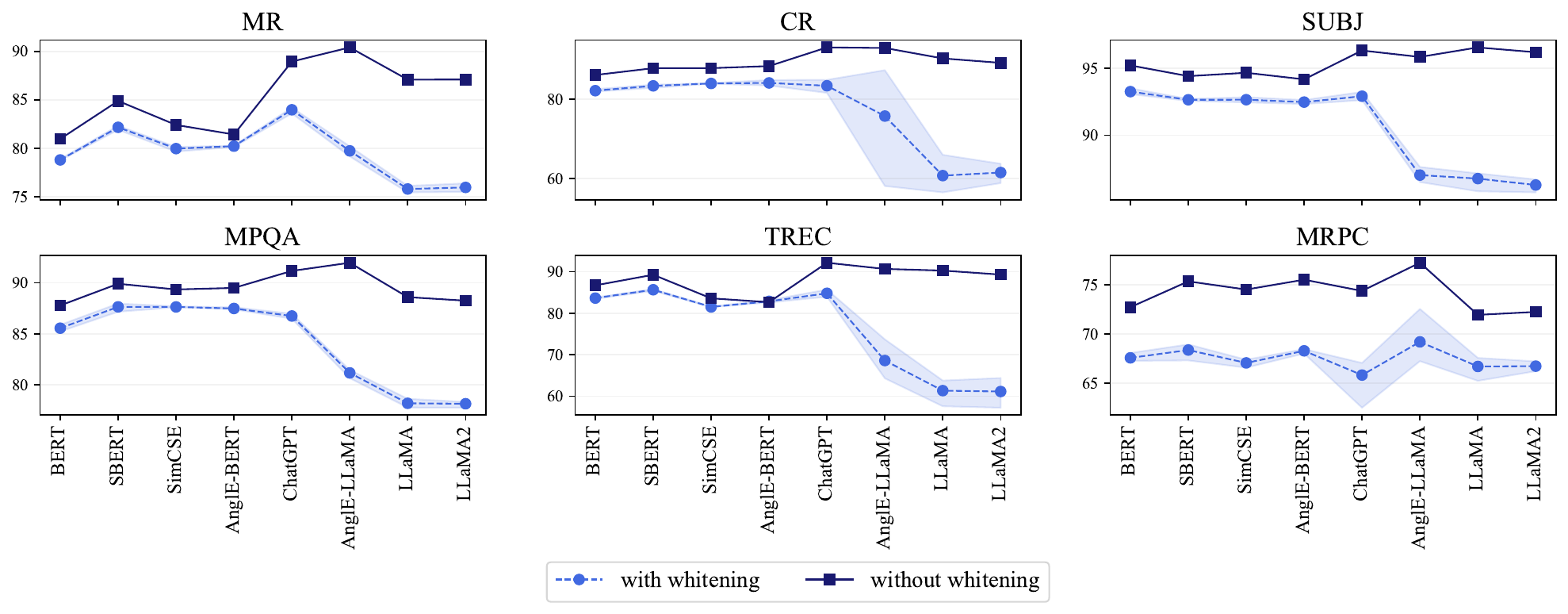}
   
    (A) Classification task. The performance is measured using accuracy per \textit{SentEval} setting because all the data sets are balanced.
    
        \includegraphics[width=0.8\textwidth]{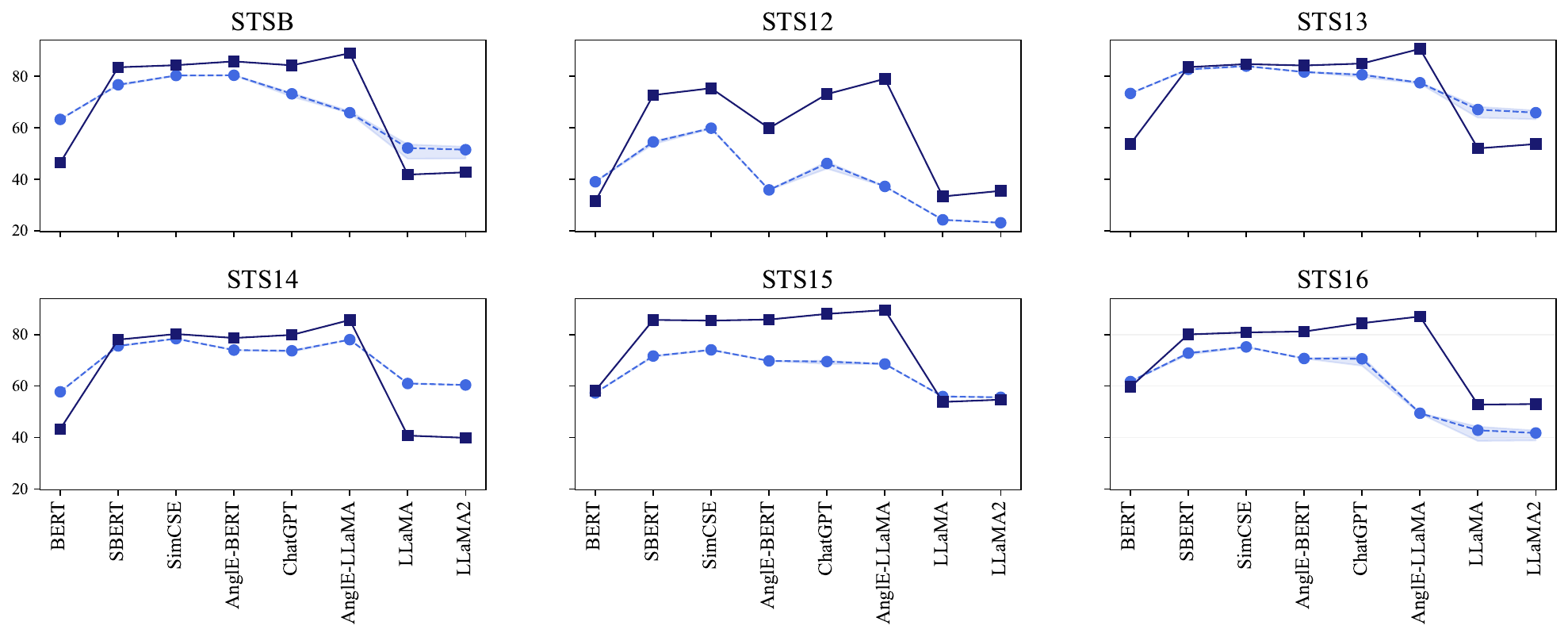}

    (B) STS task. The performance is measured using coefficient of Spearman’s correlation, expressed as a percentage. 
 
    \caption{Whitening leads to a deterioration in classification tasks (subplot A),  but demonstrates improvements in STS tasks on some models (subplot B). The performance is the average of five whitenings, with shaded area indicating the range.  }
    \label{fig:clswhitening}
\end{figure*}
\begin{table*}[hbt!]
    \centering \footnotesize
   
    \begin{tabular}{lc|ccccccc|cc}\toprule
         Model&                 		 			Dim.         & MR             & CR             & SUBJ           & MPQA            & TREC              & MRPC           & SST-F          & Avg              \\\cmidrule{3-9}
         \# Samples 			     				&				    & 10,664		 & 3,777          & 10,002		   & 10,608		     & 5,956	  		 & 1,513 	      & 8,544          &	              \\
         \midrule
         BERT \cite{devlin2018bert}&     	 	768   			    & 80.96          & 86.17          & 95.21          & 87.78           & 86.71             & 72.73          & 46.74          & 79.47            \\
         BERT$_{\text{W}}$&         					  				& 78.79          & 82.21          & 93.25          & 85.59           & 83.67             & 67.54          & 42.44          & 76.28            \\\hline
         
         SBERT \cite{reimers2019sentence}&       768     			& 84.88          & 87.89          & 94.41          & 89.91           & 89.26             & 75.35          & 50.00          & 81.67            \\
         SBERT$_{\text{W}}$&        									& 82.18          & 83.33          & 92.64          & 87.60           & 85.55             & 68.14          & 43.85          & 77.61            \\\hline
         
         SimSCE \cite{gao2021simcse}&            768     			& 82.40          & 87.90          & 94.66          & 89.35           & 83.59             & 74.52      	  & 48.26          & 80.10            \\
         SimSCE$_{\text{W}}$&       									& 79.96          & 84.04          & 92.66          & 87.63           & 81.44             & 67.16          & 43.67          & 76.65            \\\hline
        
         AnglEBERT \cite{li2023angle}&           768       			& 81.42          & 88.42          & 94.17          & 89.50           & 82.66             & 75.52       	  & 44.88          & 79.51            \\
         AnglE-BERT$_{\text{W}}$&									& 80.22          & 84.19          & 92.50          & 87.47           & 82.80             & 68.28       	  & 43.41          & 76.98            \\\hline
         
         ChatGPT \cite{OpenAI}&           		1536                & 88.94          & \textbf{93.14} & 96.32          & 91.17           & \textbf{92.15}    & 74.38          & \textbf{55.02} & \textbf{84.45}   \\
         ChatGPT$_{\text{W}}$&   							    	& 83.98          & 83.25          & 92.89          & 86.72           & 84.75             & 65.18          & 44.25          & 77.29            \\\hline
         
         AnglELLaMA \cite{li2023angle}&          4096                & \textbf{90.40} & 93.00          & 95.84          & \textbf{91.97}  & 90.66             & \textbf{77.24} & 51.98          & 84.30            \\
         AnglE-LLaMA$_{\text{W}}$& 									& 79.82          & 72.26          & 86.88          & 81.18           & 67.63             & 68.79          & 37.62          & 70.45            \\\hline
         
         LLaMA \cite{touvron2023llama}&          4096       			& 87.08          & 90.36          & \textbf{96.55} & 88.60           & 90.27             & 71.95          & 46.34          & 81.45            \\
         LLaMA$_{\text{W}}$ &    									& 75.90          & 60.80          & 86.67          & 78.15           & 60.82             & 66.81          & 34.73          & 66.24            \\\hline
         
         LLaMA2\cite{touvron2023llama2}&         4096                & 87.09          & 89.24          & 96.19          & 88.25           & 89.30             & 72.25          & 47.39          & 81.39            \\
         LLaMA2$_{\text{W}}$&     									& 76.02          & 61.33          & 86.29          & 78.26           & 60.98             & 66.82          & 35.11          & 66.40            \\
    \bottomrule
    \end{tabular}
    \caption{Classification task results of 8 models on 7 datasets in accuracy. Reported results derived from our classification experiments based on \textit{SentEval} settings. 
    All datasets are binary except SST-F, which has 5 class labels.}
    \label{tab:cls}
\end{table*}

We experimented with 8 models on classification and STS tasks.
The embeddings are  extracted  from the last layer of the BERT and LLaMA models, following the practice described in \cite{reimers2019sentence}. 
We also explored other pooling strategies and observed similar pattern. Embeddings of SBert, AnglE, and SimCSE are generated using their provided frameworks. While AnglE and SimCSE typically use the CLS pooling method to extract embeddings, which involves using the output of the 'CLS' token from the model to represent the entire input sequence, SimCSE employs the mean pooling method instead. For all mentioned models, we used the original tokenizers. For generating ChatGPT embeddings, we choose the recent \textit{text-small-3-embeddings}.

Next, we employ the \textit{SentEval} setting to evaluate the embeddings. The classification setup involves using an MLP (Multi-Layer Perceptron) classifier with no hidden layers, utilizing the \textit{RMSprop} optimizer. 
We also experimented with other classifiers including logistic regression, SVM, and Random Forests. Although the accuracy of the classification varies, the overall conclusion remains the same.  Following the practice in SentEval, we report accuracy instead of F1 because the datasets are balanced.


\subsection{Classification Task}
Table \ref{tab:cls} and subplot A of Figure \ref{fig:clswhitening} summarize our experiments on classification task.
The surprising result is that whitening transformations lead to deteriorated performance on classification tasks for all models and all the datasets without exception. What is more surprising is the large gap before and after the whitening. 
The delta can be as large as -11 in LLaMA models on the MR dataset.  The gap grows as the dimension increases--the models are sorted by their dimension in increasing order. 

\begin{figure}[]
    \centering
    \includegraphics[width=0.5\textwidth]{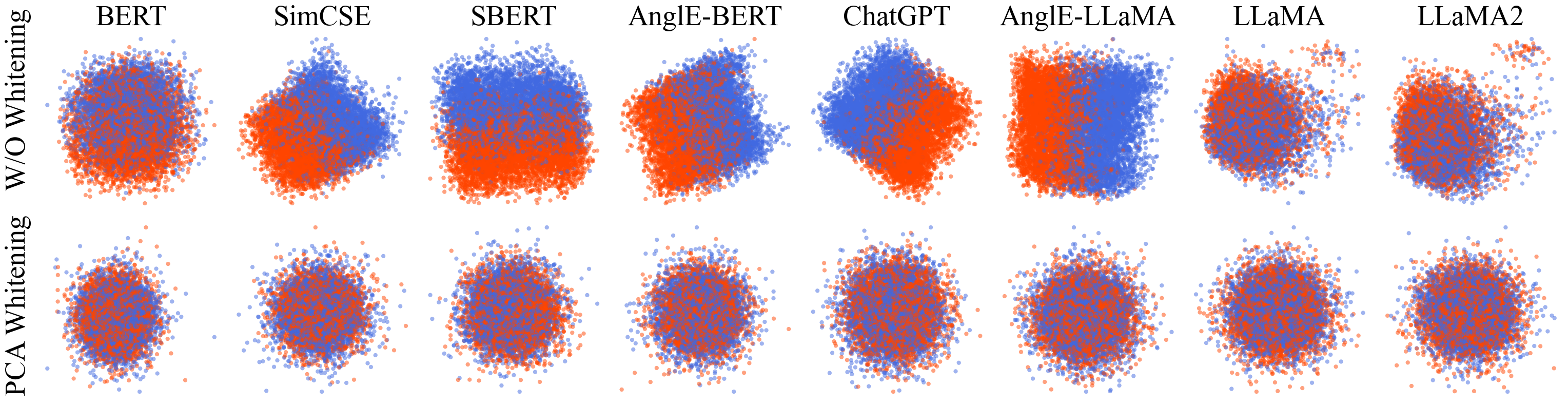}
    
    (A) 8 models embedding vs their whitenings
    
    \includegraphics[width=0.5\textwidth]{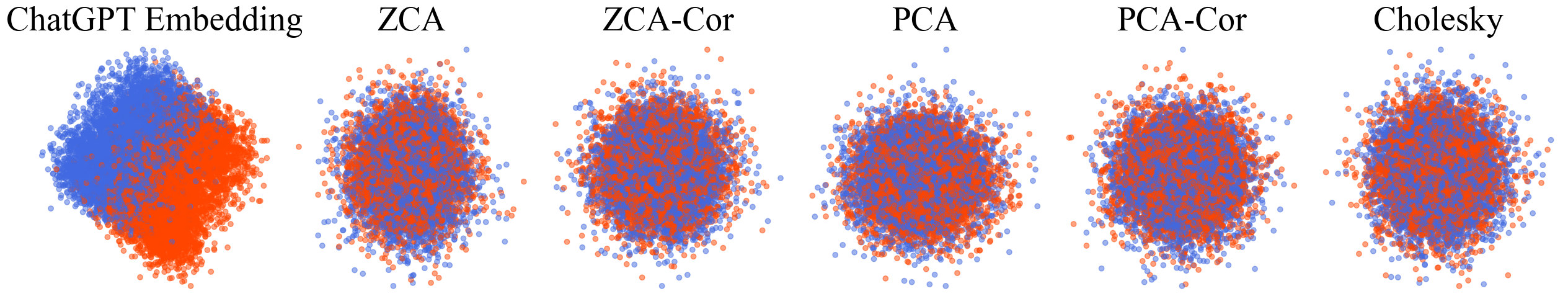}
    
    (B) ChatGPT embedding vs its five whitenings
    \caption{Visualization of embeddings before and after whitening. Dimensions are reduced using PCA. }
    \label{fig:whitenings}
\end{figure}

To understand the whitening behaviour, we visualize the embeddings before and after the whitening in Figure \ref{fig:whitenings}.  We can observe that, indeed, whitening makes features more independent but, at the same time, makes the classification more difficult. An interesting pattern is that fine-tuned models, including SimCSE, SBert, AngleBERT, and AngleLLaMA, have a distinctive square shape, while  vanilla LLaMA and BERT models do not have that pattern. That prompts us that ChatGPT may have fine-tuned their embeddings, probably using the same training data, i.e. SNLI.  

\subsection{STS Task}
Our experiments reproduced the results that are reported in \cite{su2021whitening,huang2021whiteningbert}, i.e., the whitening improves the embedding for BERT. But that conclusion can not be extrapolated to LLMs like AngleBERT, AngleLLaMA and ChatGPT. Our experiment also echoes the results from \cite{zhuo2023whitenedcse}, which shows that whitening does not work on SimCSE. Not much work has been done on the evaluation of whitening on ChatGPT and LLaMA. We find that it improves LLaMA embedding while deteriorating ChatGPT embedding.  It seems that, overall, whitening does not work for fine-tuned models.

\subsection{Impact of Whitening on Isotoropy}
\begin{figure}[]
   \centering
\includegraphics[width=0.48\textwidth]{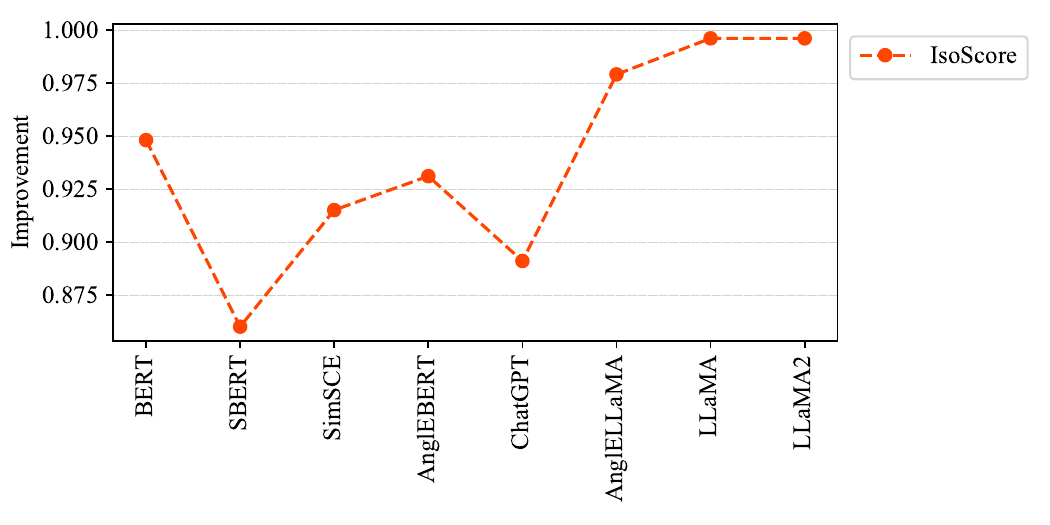}
   \caption{Improvement in Isotropy measured with IsoScore due to Whitening on MR dataset. 
   }
    \label{fig:iso}
\end{figure}
Whitening transformation ensures data isotropy by making the covariance matrix proportional to the identity matrix, thus normalizing variance across dimensions \cite{rudman2023stable,rudman2021isoscore,rajaee2021does}. Traditional isotropy metrics like average random cosine similarity score, partition isotropy score, intrinsic dimensionality, and variance explained ratio are often used in research to evaluate the isotropy of embeddings \cite{rudman2021isoscore}. However, IsoScore suggests these methods do not accurately measure isotropy. IsoScore, which applies PCA to ensure dimension independence and then assesses how the normalized variance deviates from the identity matrix, ranges from 0 to 1, indicating how uniformly data occupies the vector space \cite{rudman2021isoscore}. This makes IsoScore unique as it is mean-independent, invariant to scalar changes in the covariance matrix, and rotation-proof, offering linear scalability with dimensionality and stability across distributions with highly isotropic subspaces. Therefore, we use IsoScore to assess the isotropy of our embeddings in this study \cite{rudman2021isoscore}.

Our results demonstrate that whitening significantly reduces isotropic bias, as evidenced by the improved IsoScore depicted in Figure \ref{fig:iso}. However, enhancing isotropy does not necessarily translate to improved performance in machine learning tasks. For instance, as shown in Figure \ref{fig:iso}, the IsoScore for the LLaMA2 embeddings increased to nearly 1 following whitening. This means that initially, the LLaMA2 embeddings exhibited a very low IsoScore, close to 0, indicating severe anisotropy. After whitening, the embeddings achieved a near-perfect isotropic distribution, reflected by an IsoScore of 1. 

We also observe from Figure \ref{fig:iso} that vanilla methods, such as LLaMA and BERT, experience a higher degree of improvement in their IsoScore compared with fine-tuned models such as SBERT and SimCSE. Suggesting that the low improvement in IsoScore of ChatGPT embeddings is a result of fine-tuning on NLI datasets.

\section{Conclusion}
We show that the performance of whitening is model-dependent and task-dependent.  For classification
tasks, we do not recommend to apply whitening. For STS tasks, the performance varies from model to model. We conjecture that it works only for LLMs before fine-tuning. Also, the technical details of ChatGPT remain to be a mystery. Based on its reaction to the whitening operation, we can infer that it may be fine-tuned, probably using NLI data. 
Another contribution of our work is an embedding evaluation platform for LLMs.

\newpage
\bibliography{ACL}
\end{document}